\documentclass[letterpaper, 10 pt, conference]{ieeeconf}

\usepackage{graphicx}
\usepackage{xcolor}
\usepackage{hyperref}
\usepackage{ifsym}
\usepackage{tikz}
\usepackage{authblk}

\graphicspath{ {./images/} }

\IEEEoverridecommandlockouts
\usepackage{xcolor}
\usepackage{hyperref}

\usepackage{hyperref}
\usepackage{graphicx}

\title{\LARGE \bf
End-To-End Training and Testing Gamification Framework \\ to Learn Human Highway Driving
}

\author{Satya R. Jaladi$^{1}$, Zhimin Chen$^{1}$, Narahari R. Malayanur$^{1}$, Raja M. Macherla$^{1}$, and Bing Li$^{1}$
\thanks{The authors are with the Department of Automotive Engineering, Clemson University International Center for Automotive Research (CU-ICAR), 4 Research Dr, Greenville, SC 29607, USA. \\ Corresponding author: Dr. Bing Li, \texttt{bli4@clemson.edu}}}

\begin{document}
\maketitle

\begin{abstract}
The current autonomous stack is well modularized and consists of perception, decision making and control in a handcrafted framework. With the advances in artificial intelligence (AI) and computing resources, researchers have been pushing the development of end-to-end AI for autonomous driving, at least in problems of small searching space such as in highway scenarios, and more and more photorealistic simulation will be critical for efficient learning. In this research, we propose a novel game-based end-to-end learning and testing framework for autonomous vehicle highway driving, by learning from human driving skills. Firstly, we utilize the popular game Grand Theft Auto V (GTA V) to collect highway driving data with our proposed programmable labels. Then, an end-to-end architecture predicts the steering and throttle values that control the vehicle by the image of the game screen. The predicted control values are sent to the game via a virtual controller to keep the vehicle in lane and avoid collisions with other vehicles on the road. The proposed solution is validated in GTA V games, and the results demonstrate the effectiveness of this end-to-end gamification framework for learning human driving skills.

\end{abstract}


\section{Introduction}

American drivers as a collective cover nearly $3.2$ trillion miles every year and nearly $23$\% of the driving is done on the highways. A study also shows that Americans spend a total of $71$ billion hours every year travelling on the highways~\cite{kim2019american}. Autonomous driving on highway roads can save a lot of time for people who commute daily and avoid fatal accidents at high speeds. This also applies to vehicles such semis and heavy-duty trucks that are often involved in accidents due to driver error. Highways are also suitable for the implementation of autonomous vehicles due to their well-structured roadways and pedestrian free environment. The current autonomous driving technologies rely a lot on computer vision and AI to solve various problems such as lane detection, traffic sign and pedestrian detection etc.

Deep learning using CNN architecture over the past few years have been playing a major role in the development of computer vision technologies, specifically in the field of object detection and lane detection algorithms. However, in order to perfect these algorithms a significant amount of training and testing is required. Due to the scenario of autonomous vehicles, these cannot be tested in regular environment as it may cause danger to regular pedestrians and vehicles. Building a specific facility is very costly. Apart from this, significant amount of data is required to train these algorithms which would again lean to higher cost and time.

In order to tackle this issue, various simulated environments have been developed which reduce the cost and time required to train and test algorithms. One major problem that many simulated environments are facing is their resemblance to reality or lack thereof. Due to noticeable difference in the visually quality of the simulated world and real-world environment, the trained algorithms cannot be readily deployed into the real world. Here the open world games such as GTA V are developed by games studios with close resemblance to reality, and an intuitive way to decrease the dependency on the real-world data is to utilize data collected in human-driving games~\cite{martinez2017beyond,yun2021simulation,singh2021action, ru2022bounded, zhang2021multi, liu2024deep, chen2023self, chen2023learning, zhou2024visual}.

\section{Related works}

The approach of deep reinforcement learning has also been tested for the development for self-driving vehicles. In this case the agent (the vehicles in our case) gathers the environmental information and tries to maximize its rewards based on a defined policy~\cite{kiran2021deep, chen2023generative, zhou2023thread, xu2024rhine}. In case of self-driving cars, the agent could be rewarded for smooth driving and avoiding colliding with other vehicles and following the traffic rules~\cite{yang2021independent, zhou2023towards, liu2024particle}. The concept of reverse imitation learning has also been studied to train vehicles to  perform self driving. Here they have used a path planner algorithm to generate a set of paths and then use inverse reinforcement learning to always select the path that gives agents the highest reward~\cite{phan2022driving, wu2024new, zhou2022fine, arief2020sane, liu2024image}. In this scenario also it requires a simulated environment and large amounts of data for training and testing.

The use of convectional neural network has been tested in off-road driving scenarios by Muller~\cite{muller2005off, li2024ecnet, ding2018vehicle}. A driving root Dave is introduced which mapped input images to steering angles. In one publication~\cite{huval2015empirical}, the authors described a CNN system that detects vehicles and lane markings for highway markings, showing that CNNs are useful in self driving.

Recently studies in navigation~\cite{yun2021virtualization,bojarski2016end} have demonstrated the
potential of end-to-end mapping image to control. As the
network is required to clone the behavior of the training, this is termed as behavior cloning. Instead of using a complicated robotics approach where the vehicle should perform actions like localization, state estimation and navigation~\cite{amini2020learning,prakash2021multi, yao2023ndc, zhu2021pseudo, perez2024integrating, li2023mask}, the training of autonomous vehicle can be simplified by using the process of behavior cloning, where the neural network learns how to drive the vehicle by observing how a human or a trained agent drives the vehicle~\cite{lee2020autonomous, xu2022data}. Although this approach is not the best case for solving the problem of autonomy, this approach can be used in environments like highways and parking zones. This approach is similar to how a human learns how to drive rather than a robotics approach~\cite{kocic2019end,liu2017path, li2023hierarchical, oehmcke2024deep, li2024cpseg}. 

The approach of deep learning has also been used for segmented tasks such as obstacle detection as mentioned in~\cite{ramos2017detecting, xu2022multi}. Here they combined the use of deep learning and geometric modeling to detect obstacles in the self driving car’s environment. The primary detection is done by the deep learning part but it takes help from the cues coming from the geometric model based on a stereo camera. There has also been several studies that show the current issues and challenges faced while using deep learning to implement scene detection and scene perception~\cite{gupta2021deep,deng2023long, zhou2023semantic, huang2022semi}. The common theme in most of the issues faced by deep learning is the lack readily available data or high amount of resources required to generate that amount of data. 

The evaluation of deep learning techniques to perform object detection in different driving conditions has been studied~\cite{simhambhatla2019self, deng2023plgslam, huang2024symmetric}, here the use of transfer learning has been discussed to have an edge while training the neural networks to detect objects in different environments. Similar to the results in our approach transfer learning has produced a significant improvement while training and inference of the networks. The use of virtual data to train a deep learning network and inference it in the real world data has been tested~\cite{taylor2007ovvv, deng2024compact, deng2023prosgnerf, wang2024jointly}. The use of video games to collect data and augment existing data-sets has been studied here~\cite{shafaei2016play, deng2024neslam}. This helped them increase the size of the data, which is crucial for the training performance and also collect data that is hard to find in real life.

In this work, we investigate whether the network behavior cloning have promising results in game environment. For this purpose, we collected a data-set of GTA highway driving with our programmable labels. The data-set and our configurations will be shared to the community for future game-based driving study and autonomous driving education. Furthermore, we propose a novel game-based learning and testing framework for automated highway driving based on deep learning. Finally, the proposed framework solution is validated in GTA V games, and the results demonstration show the effectiveness of the approaches. Our main contributions are following:

\begin{enumerate}
\item{
We proposed a game-based learning and testing framework for automated highway driving based on deep learning to learn from human driving skills.}
\item{
We validated the framework solution by training and testing an end-to-end network model, and our demonstration show the effectiveness of the approaches.}
\item{
We collected a dataset of GTA highway driving with our programmable labels. The dataset and code are shared to the community for game-based driving study and autonomous driving research.}
\end{enumerate}

   \begin{figure}[h]
      \centering
      \includegraphics[width=0.5\textwidth]{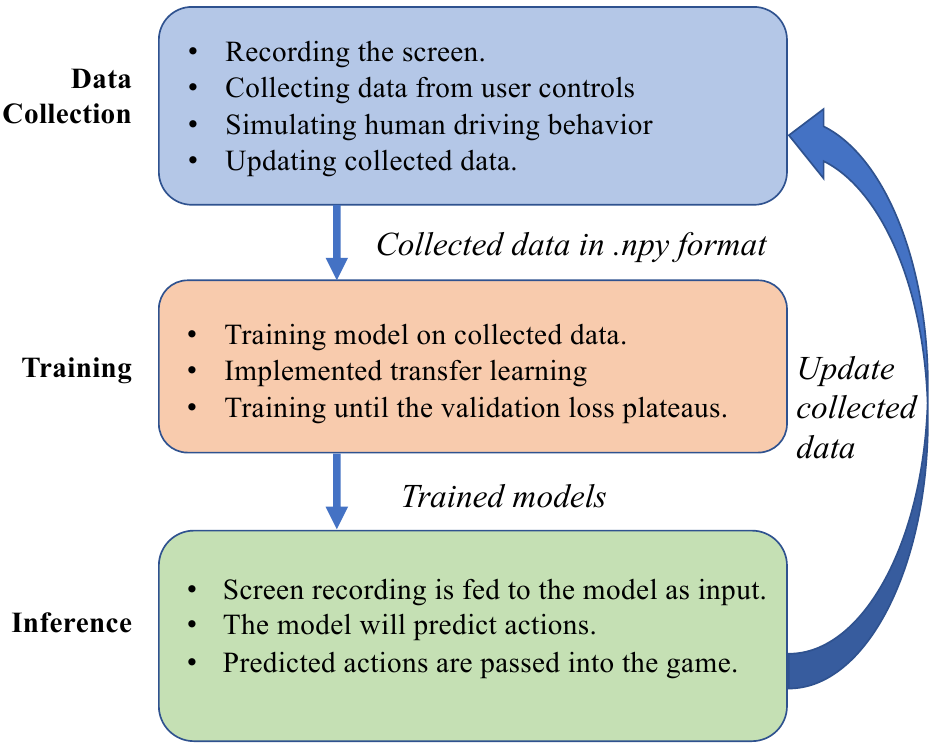}
      \caption{Framework methodology explaining the methodology implemented to perform data collection, training and inference}
      \label{figurelabel}
   \end{figure}

\vspace{+2mm}
\section{End-To-End Learning Framework}

In order to learn from human driving, we leverage existing gamification solutions such as Grand Theft Auto V (GVA V), under typical highway driving map scenarios. Two neural networks were trained using the same data and tested on a highway environment of GVA-V. The networks are trained using ‘end-to-end’ learning method as they need to take the input from one end and provide the desired output without any modification in between. 

The Nvidia’s end-to-end network architecture~\cite{bojarski2016end, tong2022adaptive} and VGG-19 architecture were used as our two neural networks of choice. As shown in  Fig. 1 the Nvidia’s architecture consists of nine layers, including a normalization layer, five convolutional layers, and three fully connected layers. The input images are converted into YUV image format before being passed through the architecture.

The initial layer of the network architecture performs image normalization to reduce the complexity of calculations and is a standard practice. The architecture is then followed by a set of strided and non-strided convolution layers. The main aspect of the convolution layers is to extract various features of varying complexity from the input image. The kernel size for these layers has been empirically chosen by a set of experiments as mentioned in~\cite{bojarski2016end, tong2024large, li2024ddnslam}. Then the architecture is followed by set of three fully connected layers. These fully connected layers  would help in converting the features extracted from the images to steering and throttle values. The final node from the original architecture has been modified from having just one node that predicts the steering value to two nodes that predict steering and throttle values as it suits out testing purpose.

VGG-19~\cite{berg2010large} has been selected to be the second network architecture to train and test on the same data.VGG-19 is primarily used to classify images and it outshined a lot of other state-of-the-art models in 2014 in the ImageNet challenge. This network architecture contains a total of $16$ CNN layers and a kernel size of 3x3 was used in all the convolutional layers, for max-pooling a window of 2x2 was used with a stride of one. Rectified Linear Activation Unit (ReLu) were as the activation function in the entire network. The network is then followed by a set of three fully connected layers and the final node consists of two nodes that output the throttle and steering values.

\begin{figure}[h]
  \centering
  \includegraphics[width=0.5\textwidth]{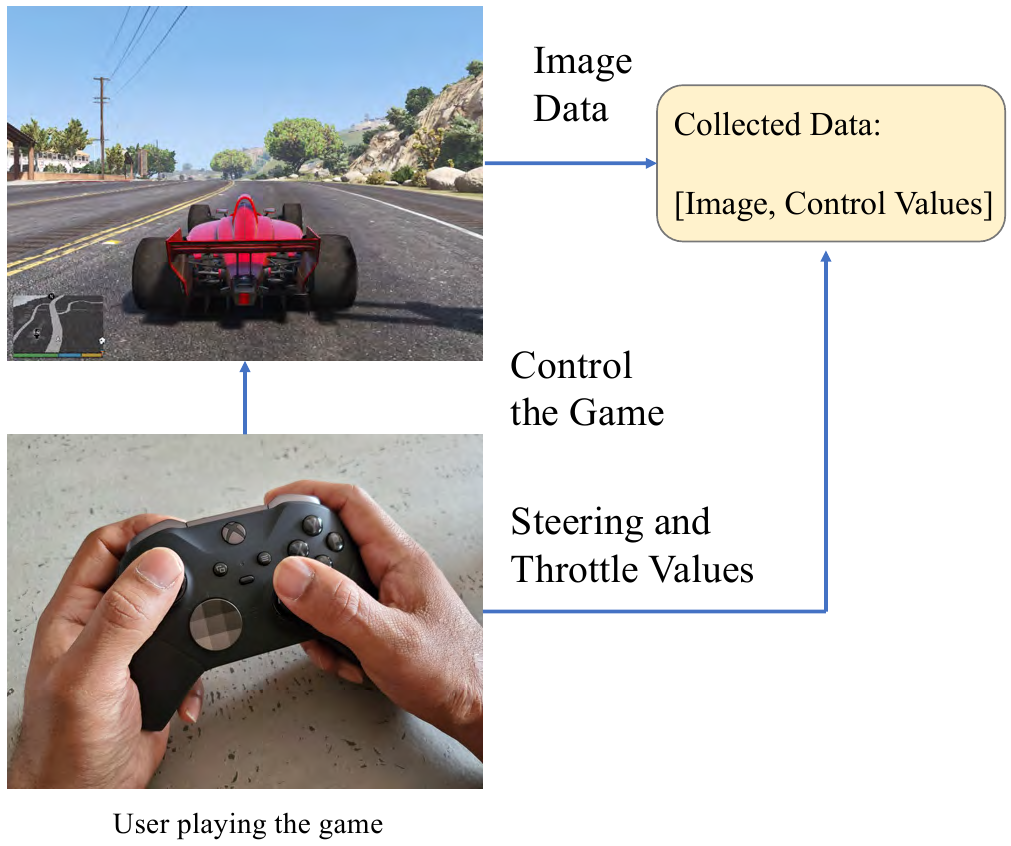}
  \caption{Data collection system (While the user plays the game, the game footage and control values are simultaneously collected)}
  \label{figurelabel}
\end{figure}

In order to avoid training the weights of the entire neural network as it takes a lot of time and resources, the approach of transfer learning has been implemented. In transfer learning approach the VGG-19 model that has been pre-trained with ImageNet dataset is imported directly. The ImageNet dataset contains around $14$ million annotated images over which the VGG-19 model has been pre-trained. 

During our training phase of this neural network, the weights for the convolution layers are frozen and only the weights of fully connected layers are trained to predict the steering and throttle values. With the help of this approach, the feature recognition aspect of the pre-trained model can be leveraged and helps us overcome the aspect of having lower amount of data to train for feature extraction. The training time required is also significantly decreased as most of the weights of the neural network are already fixed.

Both the models were imported and trained using the TensorFlow 2.0 library in Python. The TensorFlow library makes it easier to load and train the models as the models need not be built from ground-up and the complex calculations required for training are taken care by the TensorFlow functions. Both the networks will be trained on the same collected data. The training procedure and the inference results are discussed in the following sections.

\section{Experiments}

In order to collect the required data, the GTA V game is run in windowed mode. The ‘Grabscreen’ script is used to record the gameplay footage and store the images. The Xbox controller used to control the vehicle is connected to ‘Vjoy’ software which enables the ‘Pygame’ Python script to read the commands coming from the Xbox controller.

To simultaneously collect the image and throttle data, both the ‘Grabscreen’ and ‘Pygame’ scripts are included in ‘CollectData.py’ Python script. The throttle values range from $-60$ to $60$ and the steering values range from $-100$ to $100$. The values collected are stored in .npy format which will be then used to train the neural networks. The data collection system is shown in Fig. 2.

\subsection{Data Collection Strategies} 

The data collected during this process should serve two main purposes, to train the vehicle to stay in the correct lane and to avoid hitting the obstacles/vehicles in front. The data collection strategies implemented to achieve the two-fold purpose are described here. The concept of teaching a neural network how to drive based on how a human being drives is termed as imitation learning. 

In this case, an actual person is tasked to drive a vehicle in the highway environment of the game. The corresponding images from the game and control inputs from the person driving the vehicle are recorded simultaneously. The configuration of the controller software is shown in Fig. 3.

First, to teach the vehicle to stay inside the lane, close to $45$\% of the entire data is collected by driving the vehicle strictly at the center of the lane. This teaches the neural network that when the vehicle is at the center of the lane, the steering input should be close to zero to continue moving in a straight path.

\begin{figure}[h]
  \centering
  \includegraphics[width=0.5\textwidth]{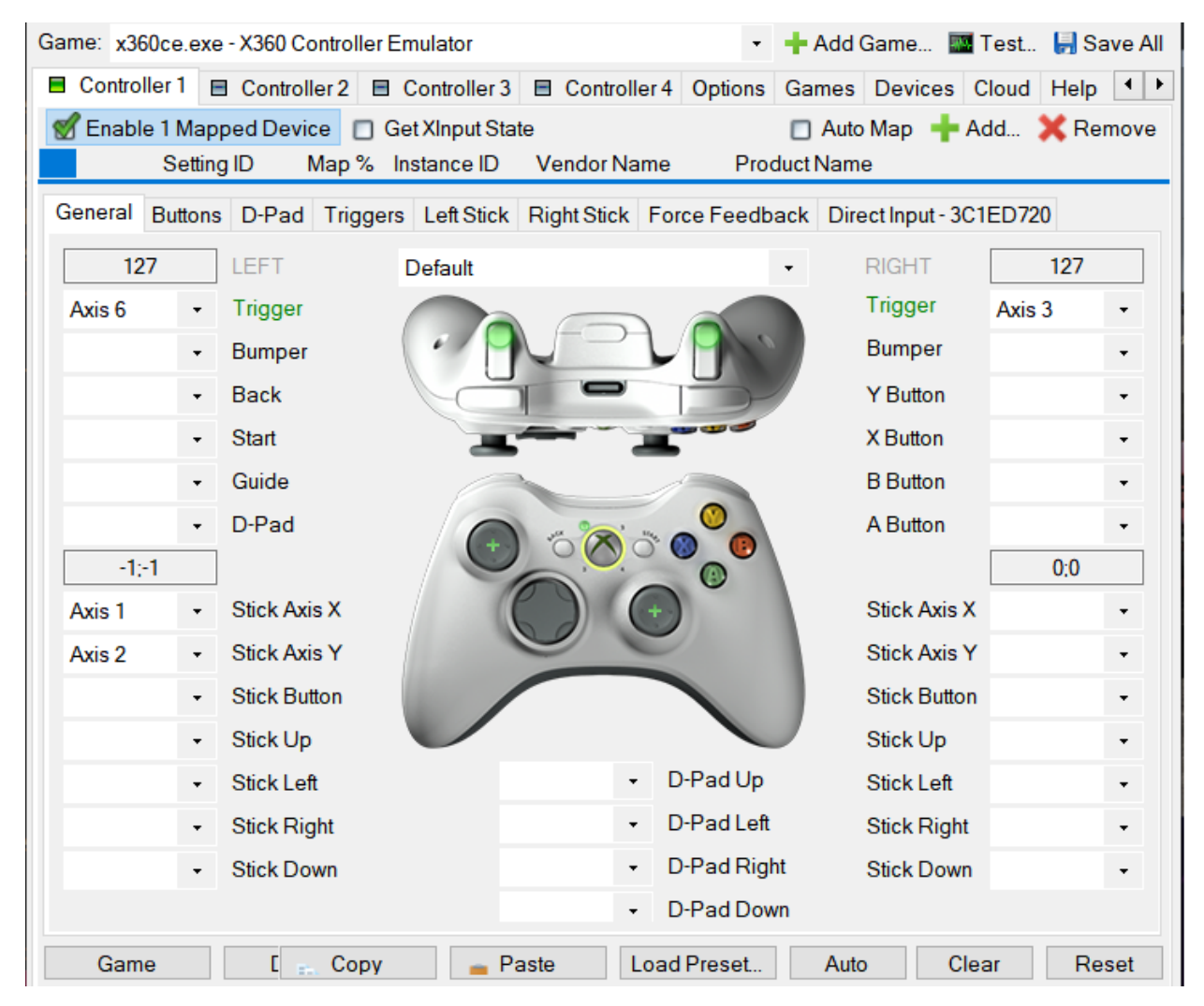}
  \caption{Virtual controller software (Used to transfer output from neural network to the in-game vehicle)}
  \label{figurelabel}
\end{figure}

The next $20$\% of the data was collected to teach the network what to do when the vehicle is about to cross a lane or deviate from the current lane. During the data collection, the user driving the vehicle lets the vehicle go out of the lane but steers it back as soon as the wheels of the vehicle start to cross the lane marking. This process is done repeatedly to collect enough data that teaches the neural network on how to bring the vehicle back to the center of the lane.

The rest of the data was collected to teach the network to apply brakes i.e., change the throttle value to negative when there’s a vehicle in front. The user driving the vehicle repeats the process of applying the brakes when the vehicle is close to another vehicle. These proportion of the three sets of data is later altered based on the inference performance. For example, if the vehicle continues to go out of lane during inference, the proportion of data teaching the vehicle to return to center of the lane is increase and tested again. 

   \begin{figure}[h]
      \centering
      \includegraphics[width=0.45\textwidth]{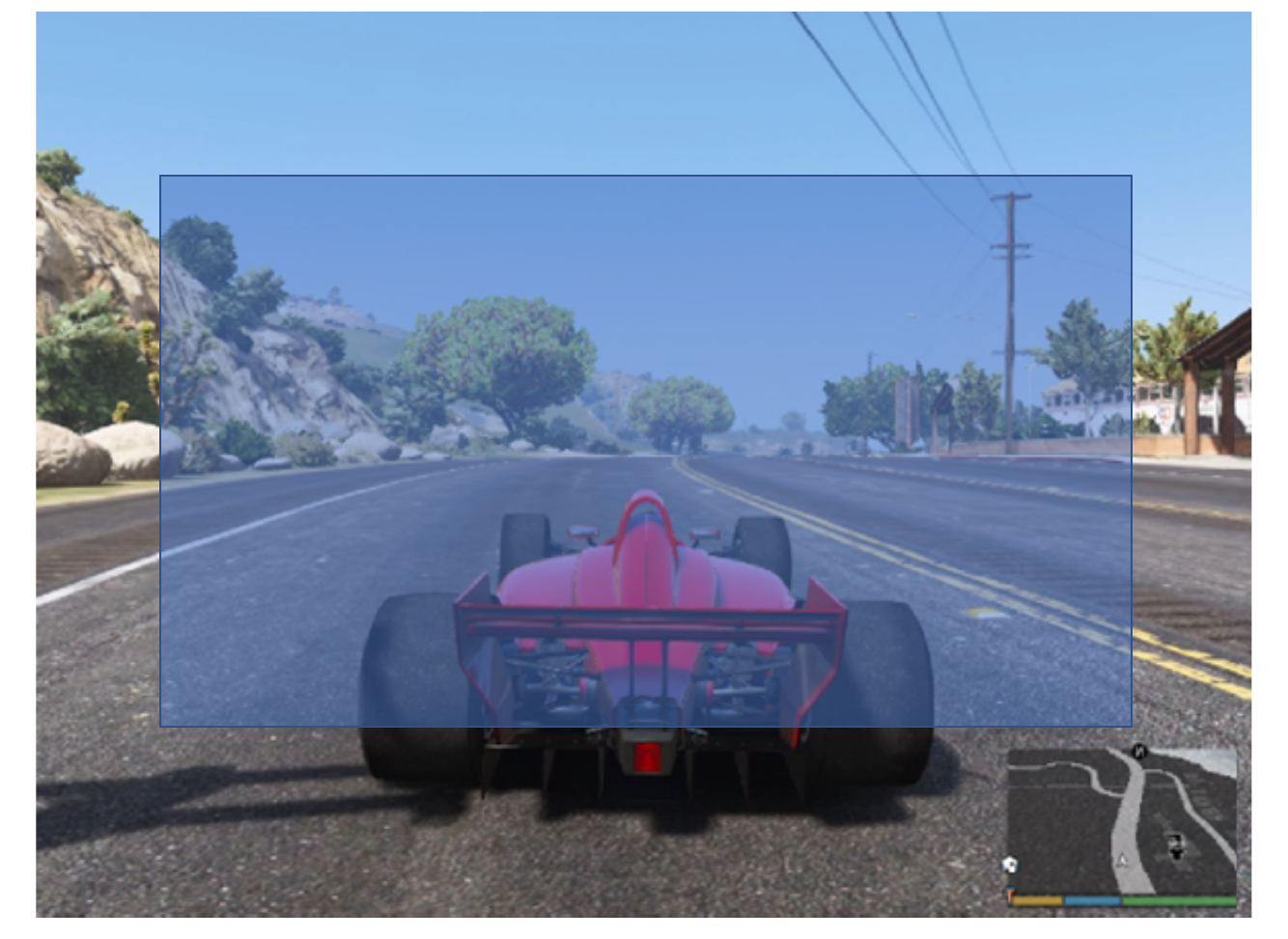}
      \caption{Image cropping (Only a certain section of the screen was fed into the algorithm to avoid redundant data)}
      \label{figurelabel}
   \end{figure}

\subsection{Image Pre-Processing and Data Augmentation}

Image pre-processing and some data augmentation techniques were implemented to clean and improve the data collected. The image was cropped as shown in the Fig. 4 such that it does not contain unnecessary portion of the image that could be the mountains and trees. Random images were selected and were flipped as shown in Fig. 5 to generalize the model and increase the size of the data collected.

\vspace{+3mm}
The brightness of the images was also adjusted to make the images darker to generalize the training for dark and bright environments. To balance the dataset which contains most of the images with forward throttle and a constant steering angle, the images with steering angle more than $0.3$ were doubled and the images with forward throttle were randomly selected and were discarded from the dataset. 

   \begin{figure}[h]
      \centering
      \includegraphics[width=0.5\textwidth]{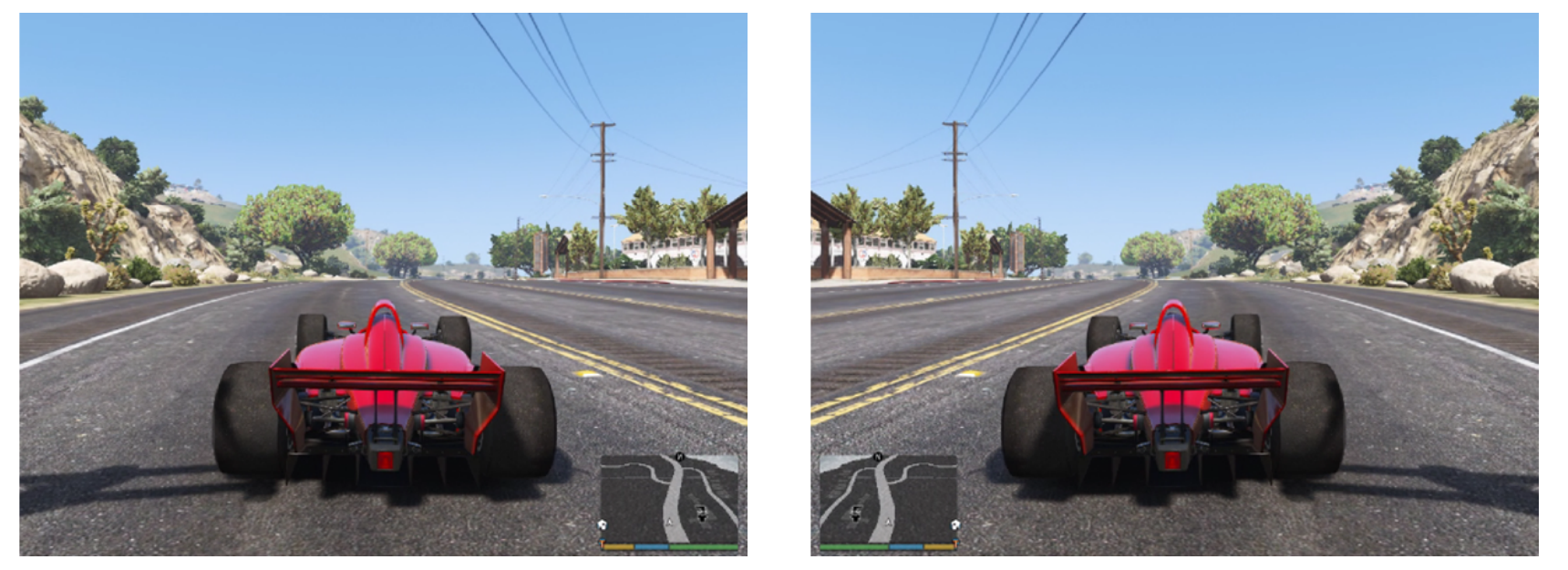}
      \caption{Image left right flipping (Augmenting data by flipping images)}
      \label{figurelabel}
   \end{figure}

\subsection{Training the Networks With Collected Data}

\vspace{+1mm}
The train dataset consisted of $70,000$ images from the GTA V game along with $15,000$ images that were used for validation. Both the models were trained with mean-square error as their error function and Adam as the optimizer. Both the proposed models were trained until the validation loss steadily decreased and care was taken to ensure the network generalized well and did not fit any of the training data. This included early stopping and other regularization techniques like data augmentation, etc. The results of the training process will be discussed in the results section. 

\vspace{+2mm}
All the models were implemented in TensorFlow 2.0 and were trained on a mobile RTX 2060 with 6GB of video memory. The optimizer used was Adam that regulates the learning rate automatically for each parameter
   
\subsection{Model Inference}

\vspace{+1mm}
Here the process of how a trained model is implemented in the game environment is demonstrated. To test either of the trained models their ability to control the vehicle in GTA V, the game was set to run in 480p resolution, in a windowed mode. As shown in Fig. 6, the Python inference script was programmed to take screenshots of the image where the GTA V game window is placed.

\vspace{+1mm}
The Grabscreen package was used to record the screen at where the GTA V window is placed and feed the images into the Python script used for inference. These images were sent to the image processing pipeline which was used for training the neural network in the first place. The images were converted to $200$x$60$-dimension length by width, converted to YUV format and normalized before feeding into the network~\cite{krizhevsky2012imagenet}.

\vspace{+3mm}
\begin{figure}[h]
  \centering
  \includegraphics[width=0.49\textwidth]{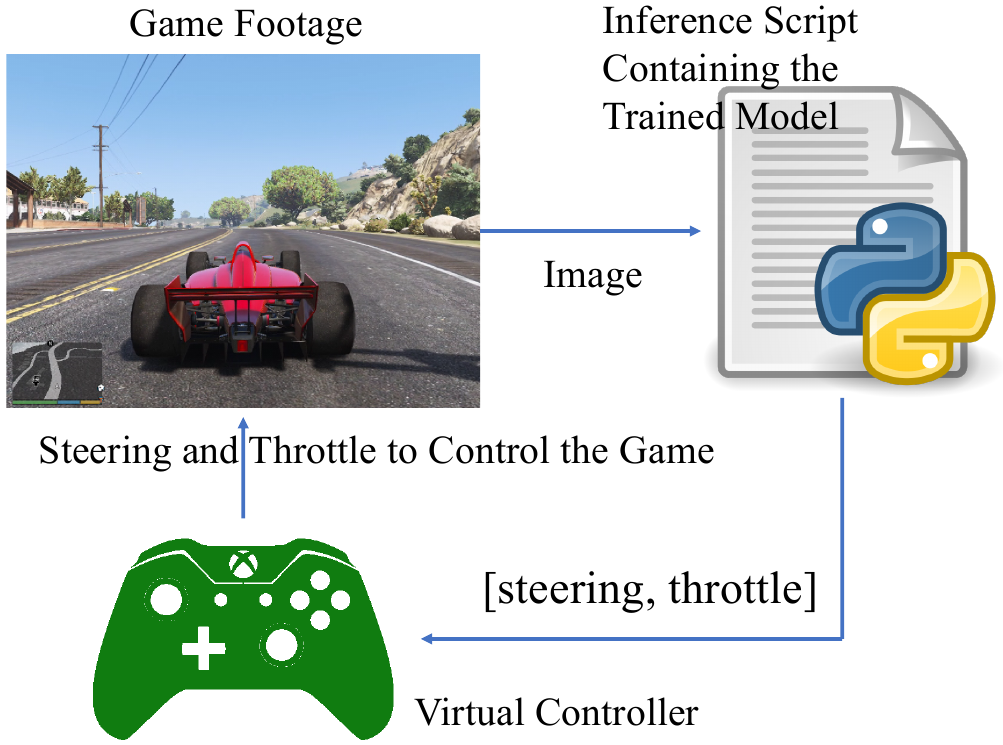}
  \vspace{+2mm}
  \caption{Inference architecture (Each image frame is sent to the Python script containing the trained architecture. The trained architecture outputs the control values via a virtual controller)}
  \label{figurelabel}
\end{figure}

\begin{figure}[h]
  \centering
  \includegraphics[width=0.5\textwidth]{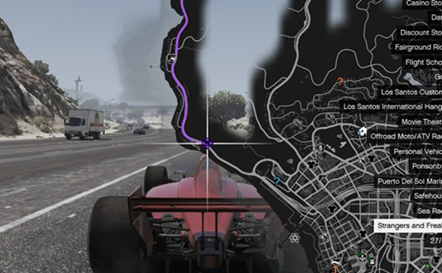}
  \vspace{+2mm}
  \caption{Location on the GTA V map where the highway data has been collected, and this location has nearly $29$ miles stretch of highway available}
  \label{figurelabel}
\end{figure}

\vspace{+2mm}
The neural network takes the image and predicts the steering and throttle values required to control the vehicle. Fig. 7 shows the exact location in the game where the inference performs.The steering control is multiplied by a gain of $1.5$ to improve the steering response and the throttle is capped to $60$\% to prevent the vehicle moving at high speed. The Vjoy package has been used to act as a virtual Xbox controller to control the vehicle in GTA V game.

\vspace{+2mm}
The package has been set up to control the vehicle in GTA V and is initialized in the inference script. The predicted values coming from the neural network are sent to the Vjoy object which directly controls the vehicle in GTA V. 

\vspace{+2mm}
The whole system acts as a closed loop control, where the predicted values are sent to control the vehicle and the changes in the game environment are returned by the Grabscreen package which in turn changes the predictions given by the neural network.

\vspace{+2mm}

\section{Results and Discussion}

\vspace{+1mm}
The two selected networks were trained on a mobile RTX 2060 GPU with 6GB of video ram. The networks were trained until the validation loss steadily decreases and the best weights were saved for inference. Out of the two models tested since the VGG-19 model was pre-trained with ImageNet dataset, it converged faster compared to the Nvidia’s architecture which were trained from scratch. The total training time for the VGG-19 network was one hour twenty five minutes whereas the Nvidia’s network took greater than six hours to train.

\begin{figure}[h]
  \centering
  \includegraphics[width=0.43\textwidth]{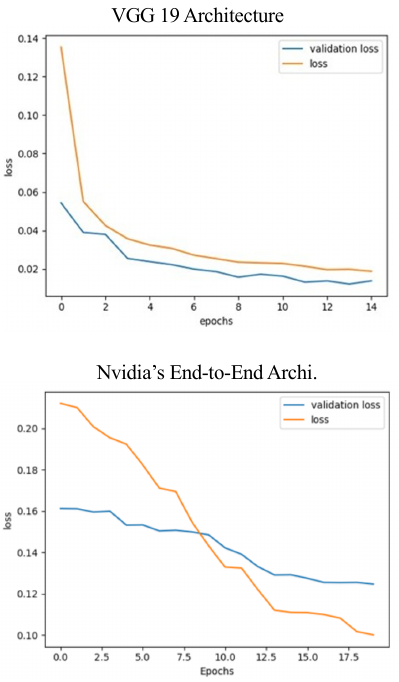}
  \caption{Training performance comparison of the two model architectures}
  \label{figurelabel}
\end{figure}

\vspace{+3mm}
As shown in Fig. 8, the VGG-19 provided the least amount of validation loss. The models were trained for $20$ epochs with a batch size of $128$. The training results show that the VGG-19’s validation loss is consistently lower than training loss, which implies the model generalized well and did not suffer from overfitting. However, the validation loss for Nvidia’s architecture did go higher than the training loss. Compared to VGG-19, it also has higher validation loss at $0.12$. Both the models were also implemented to drive the vehicle inside the game the performance is measured.

\vspace{+3mm}
The trained VGG-19 architecture provided good performance in controlling the vehicle by keeping the vehicle in lane and avoiding the vehicle from colliding with the other vehicle in the highway environment. 

\vspace{+3mm}
Due to the constraints of running both the video game and the inference on the same mobile RTX 2060 GPU, the model is only able to process between $5$-$10$ frames per second. Even though this could negatively affect the performance and response time of the network, it did perform considerably well.

\begin{figure}[h]
  \centering
  \includegraphics[width=0.5\textwidth]{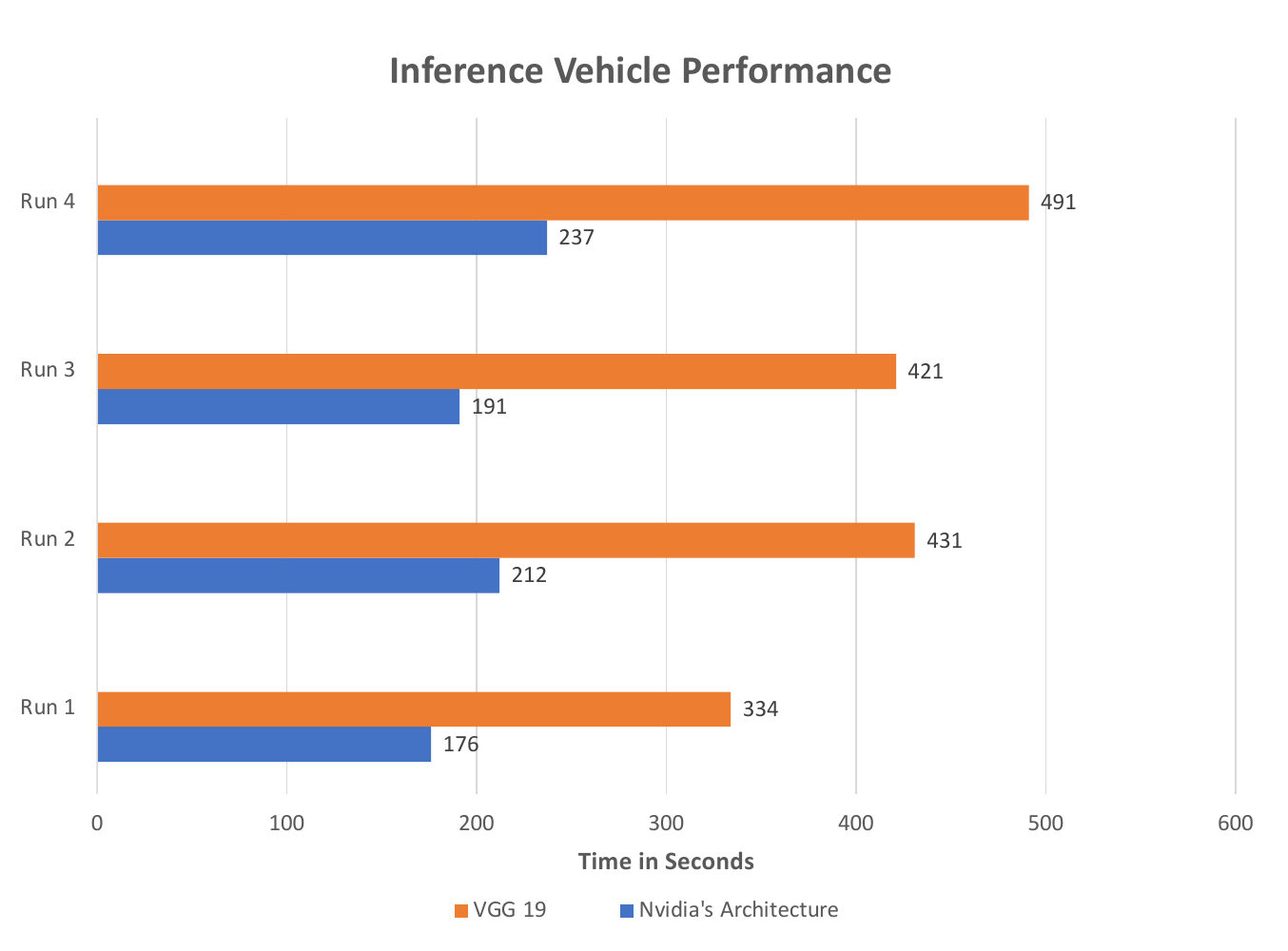}
  \vspace{+2mm}
  \caption{Inference results (showing how long the vehicle traveled inside the game without completely going out of the lane or hitting another vehicle.)}
  \label{figurelabel}
\end{figure}

\vspace{+3mm}
Fig. 9 shows the time driven by the vehicle without completely going out of lane or hitting another vehicle, and the VGG-19 model consistently performed better than the Nvidia’s architecture. After each run the training data collected is modified and the models are re-trained again for better performance, hence the overall performance of both the models increased after every run.

\vspace{+3mm}
As this is an end-to-end architecture, the actions of the vehicle performed during inference cannot be strictly justified, for example we cannot particularly explain why the vehicle was going out of lane in certain scenarios, it is a case of assuming what would have caused the behavior and improvising the training data to negate such undesired instances. In the real world scenarios, CNN networks are primarily used in a modular methodology where their job would only be to detect the lane or obstacles near a vehicle and rest of the control algorithm would be taken care by another control algorithm. Our research result code \footnote{\href{https://github.com/AutoAILab/End2EndDriving}{\textcolor{blue}{https://github.com/AutoAILab/End2EndDriving}}} and comprehensive demonstration video \footnote{\href{https://www.youtube.com/watch?v=tUFN5y6eTRo}{\textcolor{blue}{https://www.youtube.com/watch?v=tUFN5y6eTRo}}} can be seen online. 

\section{Conclusion}

An end-to-end training and testing gamification framework are proposed in this research, with a CNN model deployed in our framework to learn the human-driving skills in GTA V environment. The vehicle was trained to stay in lane and avoid collisions with the vehicle around it. The performance of the network has been satisfactory. Due to the limits of the GPU performance, the model was limited to running at 
five frames per second which was one of the reasons why the vehicle was running at slow response time. The future work of this research is to quantify the performance with human subject study in different scenarios with different users and create metric of driving skills learning.

\bibliography{IEEEfull}
\bibliographystyle{IEEEtran}

\end{document}